\documentclass[conference]{IEEEtran}

\IEEEoverridecommandlockouts
\usepackage{cite}
\usepackage{amsmath,amssymb,amsfonts}
\usepackage{eqnarray, algorithmic}
\usepackage{graphicx}
\usepackage{textcomp}
\usepackage{svg, xcolor, xspace}
\usepackage{float, balance}
\usepackage{listings}
\usepackage{censor}
\usepackage{paralist}
\usepackage{booktabs}
\usepackage{tabularx}
\usepackage{soul}

\usepackage[loadshadowlibrary]{todonotes}
\usepackage{url} 
\AtBeginDocument{
  \definecolor{pdfurlcolor}{rgb}{0,0,0.6}
  \definecolor{pdfcitecolor}{rgb}{0,0.6,0}
  \definecolor{pdflinkcolor}{rgb}{0.6,0,0}
  \definecolor{light}{gray}{.85}
  \definecolor{vlight}{gray}{.95}
}
\usepackage[colorlinks=true,citecolor=pdfcitecolor,urlcolor=pdfurlcolor,linkcolor=pdflinkcolor,pdfborder={0 0 0}]{hyperref}

\renewcommand{\arraystretch}{1.1}

\newcommand\BibTeX{{\rmfamily B\kern-.05em \textsc{i\kern-.025em b}\kern-.08em
T\kern-.1667em\lower.7ex\hbox{E}\kern-.125emX}}

\newcommand{\eg}{e.g.,\xspace}

\definecolor{mRed}{rgb}{0.8,0,0}
\definecolor{mGreen}{rgb}{0,0.6,0}
\definecolor{mGray}{rgb}{0.5,0.5,0.5}
\definecolor{mPurple}{rgb}{0.58,0,0.82}
\definecolor{backgroundColour}{rgb}{0.98,0.98,0.95}

\lstdefinestyle{CStyle}{
    xleftmargin=.05\linewidth, 
    linewidth=\linewidth, 
    backgroundcolor=\color{backgroundColour},
    commentstyle=\color{mGreen}\bf,
    keywordstyle=\color{magenta},
    numberstyle=\tiny\color{mGray},
    stringstyle=\color{mPurple},
    basicstyle=\ttfamily\scriptsize,
    breakatwhitespace=false,
    breaklines=true,
    captionpos=b,
    keepspaces=true,
    title=\lstname,
    numbers=left,
    numbersep=5pt,
    showspaces=false,
    showstringspaces=false,
    showtabs=false,
    tabsize=2,
    language=C++
}

\usepackage{orcidlink}
\makeatletter
\newcommand{\linebreakand}{%
  \end{@IEEEauthorhalign}
  \hfill\mbox{}\par
  \vspace{-.7\baselineskip}
  \mbox{}\hfill\begin{@IEEEauthorhalign}
}

\makeatother

\begin{document}

\title{An Agentic AI Framework to Accelerate\\ Scientific Discovery in Plant Phenotyping
\thanks{
This manuscript has been authored in part by UT-Battelle, LLC, under contract DE-AC05-00OR22725 with the US Department of Energy (DOE). The publisher, by accepting the article for publication, acknowledges that the U.S. Government retains a non-exclusive, paid up, irrevocable, world-wide license to publish or reproduce the published form of the manuscript, or allow others to do so, for U.S. Government purposes. The DOE will provide public access to these results in accordance with the DOE Public Access Plan (http://energy.gov/downloads/doe-public-access-plan).}}

\author{
    \IEEEauthorblockN{Renan Souza\,\orcidlink{0000-0002-1794-808X}}
    \and
    \IEEEauthorblockN{Daniel Rosendo\,\orcidlink{0000-0003-1175-8426}}
    \and
    \IEEEauthorblockN{Kelsey Carter\,\orcidlink{0000-0001-8327-6413}}
    \and
    \IEEEauthorblockN{John Lagergren\,\orcidlink{0000-0002-8092-7433}}
    \linebreakand
    \IEEEauthorblockN{Frédéric Suter\,\orcidlink{0000-0003-1902-1955}}
    \and
    \IEEEauthorblockN{Shelaine L. Curd\,\orcidlink{0009-0004-6419-7479}}
    \and
    \IEEEauthorblockN{Gerald A. Tuskan\,\orcidlink{0000-0003-0106-1289}}
    \and
    \IEEEauthorblockN{Rafael Ferreira da Silva\,\orcidlink{0000-0002-1720-0928}}
    \and
    \IEEEauthorblockN{David Weston\,\orcidlink{0000-0002-4794-9913}}
    \linebreakand
    \IEEEauthorblockA{
       Oak Ridge National Laboratory, Oak Ridge, TN, USA \\
       \{souzar, rosendod, carterkr, lagergrenjh, suterf, curdsl, tuskanga, silvarf, westondj\}@ornl.gov
    }
}

\maketitle

\begin{abstract}
High-throughput plant phenotyping now generates image derived datasets far faster than scientists can analyze them. At Oak Ridge National Laboratory's Advanced Plant Phenotyping Laboratory (APPL), automated stations image hundreds of plants daily across multiple remote sensing modalities; yet, trait extraction and interpretation remain manual, expert-bound, and strictly post-hoc, making analysis, not acquisition, the binding constraint on discovery.
We present an end-to-end agentic AI framework that turns the facility from a data factory into an interactive autonomous, discovery platform, where scientists partner with AI agents to accelerate time to insight.
A conversational Co-Scientist Agent translates a scientist's natural-language question into a structured analysis plan, and a headless Compute Agent dispatches Vision Transformer segmentation and trait extraction on the Frontier exascale supercomputer. The two agents run in separate security and resource domains and communicate over a secure, token-authenticated streaming channel, a design that accounts for the federation, data-movement, and provenance realities cloud-native agentic frameworks ignore, ensuring end-to-end provenance is captured for every interaction. The framework turns a days- to weeks-long analysis process into an interactive loop where agents reason over results, recommend next analyses, and respond to follow-up questions in seconds.
\end{abstract}

\section{Introduction}

Scientific discovery is increasingly constrained by the human cognitive and computational effort required to rapidly transform huge amounts of data into insight.
Recent advances in artificial intelligence (AI) promise to relax this constraint across scientific domains, and 
nations and institutions are rapidly mobilizing AI capabilities to gain or retain a competitive advantage in scientific discovery. Pressing global challenges (\eg energy dominance, advanced nuclear and fusion technology, critical minerals and materials, biomanufacturing, and national security) demand faster innovation cycles than conventional research methodologies can provide. In response, the U.S. Department of Energy (DOE) launched the Genesis Mission, a historic national effort mobilizing the 17 DOE national laboratories, industry, and academia to harness AI and high-performance computing (HPC) to fundamentally transform how American science is conducted. Anchored by an integrated American Science and Security Platform that connects the nation's leading HPC resources, AI systems, and scientific instruments, the Genesis Mission's overarching objective is to double the productivity and impact of U.S. research and development within a decade while securing American technological leadership, accelerating breakthroughs, and strengthening national security.

 Within this broader effort, biological and environmental research has traditionally benefited from advances in automation, high-throughput experimentation, and computational modeling~\cite{tom2024self}.
 Autonomous, self-driving laboratories have already demonstrated dramatic acceleration in adjacent domains such as inorganic-materials synthesis~\cite{szymanski2023autonomous}, but the complexity and nonlinearity of biological systems make them difficult to model and predict computationally, and the time-intensive nature of conventional wet-lab experimentation limits the pace at which hypotheses can be tested and validated. Moreover, critical applications such as bioengineering organisms to produce sustainable fuels, phytomining rare earth elements, or developing resilient crops for food security, require accelerated discovery cycles that conventional methodologies cannot support~\cite{wang2025sensors}. The Orchestrated Platform for Autonomous Laboratories\footnote{ Orchestrated Platform for Autonomous Laboratories: \url{https://opal-doe.org}} (OPAL) thus focuses specifically on advancing biological discovery by orchestrating a federated network of autonomous laboratories across multiple DOE national laboratories~\cite{ferreira2025grassroots} that combine AI, HPC, robotics, and automated experimentation to enable self-driving science. Its primary objective is to transform how biological research is conducted by enabling laboratories to learn, adapt, and operate continuously to dramatically accelerate breakthroughs, while maintaining human oversight and guidance of the discovery process.

\begin{figure*}[t!]
    \centering
    \includegraphics[width=\linewidth]{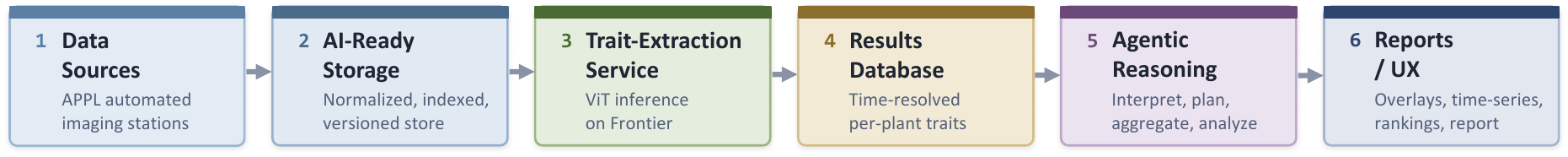}
    \caption{System architecture. Raw multimodal imagery flows left to right through AI-ready storage, on-demand ViT trait extraction on Frontier, and a results database into an agentic reasoning layer that delivers reports and visualizations. The return path forms the core loop: follow-up questions reuse established context and computed traits. Bands list the data, compute, agent, and UX layers and their responsibilities.}
    \label{fig:system}
\end{figure*}

In this paper, we make the following contributions:
\begin{compactenum}
    \item An end-to-end \textbf{AI agentic framework for high-throughput plant phenotyping} that combines HPC-scale perception models, federated multi-agent orchestration, and reasoning in a production DOE national-laboratory facility. 
    Unlike cloud-only agentic systems, the design integrates cloud-hosted services, including MCP servers and LLM servers, with Slurm-scheduled HPC workflows running on a supercomputer, and treats provenance as a first-class output rather than an afterthought.

    \item A scalable \textbf{AI-readiness pipeline} and a Vision Transformer (ViT)-based segmentation and trait-extraction service, trained on annotated multimodal imagery and deployed on the Frontier exascale supercomputer to handle multiple imaging modalities and terabyte-per-day data rates produced by the APPL facility.

    \item \textbf{A multi-agent architecture} in which a conversational Co-Scientist Agent, exposed to scientists through a web-based chat\,+\,dashboard interface, translates natural-language scientific questions into structured analysis plans, while a
  Compute Agent executes these plans by dispatching inference and trait extraction on Frontier.

    \item \textbf{End-to-end workflow provenance} through a system that captures experiment identifiers, inference and trait extraction algorithm parameters, implementation versions, model versions, LLM requests including prompts, intermediate artifacts, and agent decisions, and includes an MCP server that enables the Co-Scientist agent to delegate natural-language provenance questions by translating them into structured database queries and summarizing result sets, supporting reproducibility, auditability, and the curation of trustworthy training data for downstream biology foundation models.
\end{compactenum}

Fig.~\ref{fig:system} gives an overview of the end-to-end framework. Multimodal imagery enters on the left and is transformed, through AI-ready storage and on-demand trait extraction on Frontier, into segmentation overlays, trait time-series, and an interpretable report produced by an agentic reasoning layer.

This paper is organized as follows. Section~\ref{sec:background} provides some background on the Advanced Plant Phenotyping Laboratory and the traditional analysis workflow it has historically supported, and motivates the development of an AI agentic framework to accelerate scientific discovery. Section~\ref{sec:system} presents a high-level overview of the system, tracing the path from raw multimodal images to scientific insight via agentic workflows. Section~\ref{sec:images_to_traits} details the AI-readiness processing pipeline and the use of AI models to extract time-resolved plant traits from multimodal images. Section~\ref{sec:implem_deploy} presents our agentic framework and provides implementation and deployment details. Finally, we summarize our contributions and present future  work directions in Section~\ref{sec:ccl}.

\section{Background and Problem Setting}
\label{sec:background}

\subsection{The Advanced Plant Phenotyping Laboratory}
\label{sec:APPL-science}
The Advanced Plant Phenotyping Laboratory\footnote{Advanced Plant Phenotyping Laboratory: \url{https://www.ornl.gov/appl}}~(APPL) at Oak Ridge National Laboratory~(ORNL) operates as a high-throughput imaging environment designed to link plant genotypes to phenotypes at an unprecedented scale and speed. The facility's automated conveyor system propels up to 520 large plants per cycle along a 700-foot track through five aboveground imaging stations, operating continuously 24/7 throughout experimental campaigns, so that a multi-week campaign images as many as 10,400 plants in total. Each imaging station deploys complementary modalities, including two RGB cameras for morphological assessment, two hyperspectral sensors capturing hundreds of spectral bands beyond visible light, thermal imaging for transpiration and water-use measurements, multispectral imaging for plant water content, chlorophyll and multispectral fluorescence for photosynthetic efficiency, and 3D laser scanning for plant architecture. During typical experiments spanning two to four weeks, the system captures data for hundreds of plants multiple times, generating terabytes of multi-modal imagery per experiment. Recent additions to APPL  enable belowground root imaging via rhizoboxes equipped with RGB and near-infrared cameras, bringing the total to eight imaging modality groups: above-ground RGB, hyperspectral, thermal, multispectral, fluorescence, and 3D laser, and below-ground RGB and near-infrared.

Representative scientific workflows at APPL involve genome-wide association studies (GWAS)~\cite{uffelmann2021genome} linking genetic variants to performance traits in bioenergy crops such as poplar ({\it Populus trichocarpa})~\cite{lagergren2023few}, pennycress ({\it Thlaspi arvense}), and switchgrass ({\it Panicum virgatum}). Researchers using APPL  typically aim to identify genes conferring drought resistance, heat tolerance, or nutrient-use efficiency, and to validate the phenotypic effects of targeted modifications intended to improve photosynthetic capacity, biomass yield, critical minerals and materials capture, or biofuel conversion efficiency. Yet a critical bottleneck emerges in the post-acquisition phase: extracting biologically meaningful traits from massive, multi-modal datasets and interpreting temporal dynamics across diverse genotypes remains computationally expensive and cognitively demanding. A representative question a biologist might ask is: which plant genotypes exhibited the greatest height gain over the previous two weeks? Answering it requires integrating time-series measurements across multiple imaging modalities, accounting for environmental covariates, and associating subtle phenotypic shifts with genetic or treatment factors. Such workflows historically demand extensive manual curation and domain expertise, creating an analytical throughput bottleneck despite APPL's high imaging capacity.

Addressing this bottleneck is a core objective of the OPAL project, which aims to shift scientific practice toward AI-accelerated interactive experimentation: scientists query experimental data and receive actionable insights within minutes rather than days, refining hypotheses during active experiments rather than after. Complete, audit-grade provenance is equally essential, both to support trust and reproducibility in individual studies and to curate the high-quality, traceable corpora needed to train and fine-tune Genesis Mission's foundation models. 

\subsection{Traditional Analysis Workflow}
\label{sec:workflow}
\begin{figure}[ht!]
    \centering
    \includegraphics[width=0.46\linewidth]{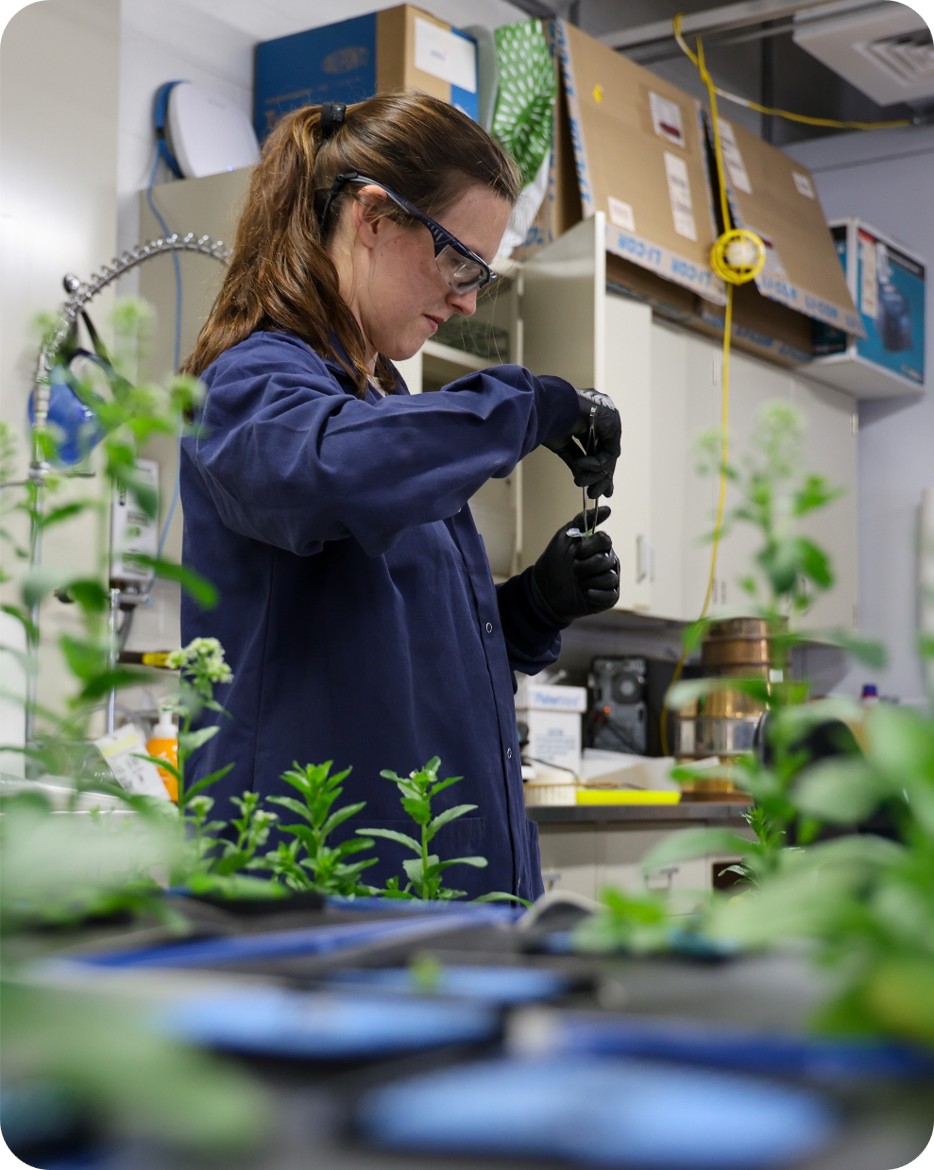}
    \includegraphics[width=0.46\linewidth]{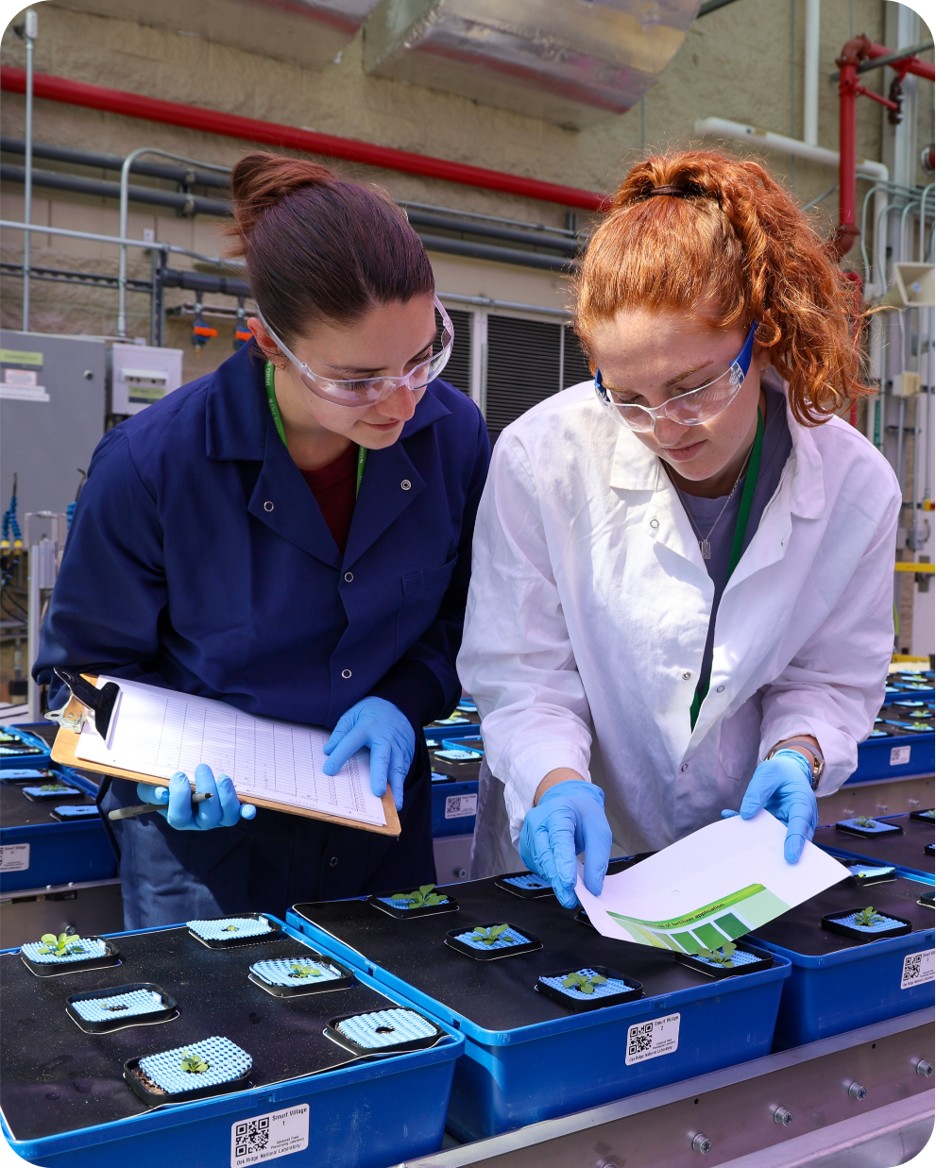}
    \caption{Sample preparation and greeness benchmarking and data collection at APPL (credit; M. Bluedorn, ORNL).}
    \label{fig:appl_preparing_benchmarking}
\end{figure}

Before automated trait extraction, characterizing plant phenotypes was an inherently manual and time-intensive process. Researchers physically handled individual plants, recorded observations by hand, and coordinated sample preparation and imaging across multiple experimental stages, as illustrated in Fig.~\ref{fig:appl_preparing_benchmarking}. Even as APPL's automated imaging stations relieved the burden of image acquisition, the downstream analysis remained largely human-driven.

APPL's automated imaging stations produce raw imagery that is deposited continuously to a local storage server, where it is checksummed, indexed against a per-experiment greenhouse database, and made available for downstream analysis. Once an imaging campaign concludes, or at periodic batch points during the campaign, the lead biologist or a designated analyst extracts a subset of the data and runs it through one of a small number of established image-analysis pipelines. Extracted traits are written to spreadsheets or per-plant trait tables that feed into downstream genetic analyses, in particular the genome-wide association studies that link APPL phenotypes back to genotype data for the plant species of interest.

Each of these pipelines is well engineered for the use case it was built around, but together they impose three structural costs. First, they are brittle across modalities and species: a pipeline tuned to segment poplar leaves under controlled RGB lighting does not generalize to hyperspectral data cubes of roughly 0.5 gigabytes each, to 3D laser-scan point clouds, or to root systems imaged through rhizoboxes, each of which requires its own segmentation strategy and its own validation effort. Second, they are programmer-bound: tuning thresholds, selecting the right color space, defining regions of interest, and choosing morphological operations all require image-analysis expertise that biologists, the actual end-users of the data, typically lack. The practical consequence is a hand-off culture in which a biologist requests an analysis, an analyst configures a pipeline, and results return days or weeks later, often after multiple iterations on edge cases. Third, the workflow is strictly post-hoc: by the time traits are extracted and analyzed, the experiment is over. There is no opportunity to use mid-experiment results to ask a follow-up question, to prioritize a subset of plants for richer phenotyping, or to steer the next imaging cycle. Multimodal integration, where it happens at all, is a final spreadsheet-level join rather than a property of the analysis itself.

The combined effect is that image analysis, not image acquisition, has become the binding constraint on what APPL can deliver. This is consistent with a broader observation across the plant phenotyping community: as automated facilities have driven imaging throughput far beyond what manual measurement could ever provide, analysis has emerged as the new bottleneck~\cite{minervini2015image}. APPL exemplifies the problem in its strongest form. A facility that can process as many as 10,400 plants per campaign around the clock across multiple imaging modalities is now limited by a downstream loop that is single-modality in practice, single-trait at a time, expert-bound, and offline. Closing this gap and doing so in a way that preserves the quality and reproducibility expected by the scientific community is the motivation for the agentic framework presented in the remainder of this paper.

\section{System Overview: From Images to Insights via Agentic Workflows}
\label{sec:system}

The pipeline of Fig.~\ref{fig:system} is organized around three successive concerns: data AI-readiness, trait extraction, and agentic reasoning. Raw multimodal images from APPL's automated imaging stations enter on the left, pass through an AI-readiness stage that normalizes and indexes them into a consistent, versioned representation, and feed a ViT-based segmentation and trait-extraction service on Frontier that writes time-resolved, per-plant traits to the OPAL data lakehouse. Together, these stages form a data and compute substrate that the right half of the pipeline builds on. On top of it sits an agentic reasoning and reporting layer that interprets natural-language requests and delivers results as segmentation overlays, trait time-series, ranked genotype lists, and a written biological interpretation. From a user's perspective, this pipeline operates as a four-stage interactive loop rather than a one-shot batch job:
\begin{table*}[t]
  \centering
  \caption{The four architectural layers and the distinct responsibility each owns.}
  \label{tab:layers}
  \footnotesize
  \renewcommand{\arraystretch}{1.25}
  \newcolumntype{Y}{>{\hsize=.6\hsize\raggedright\arraybackslash}X}
  \newcolumntype{Z}{>{\hsize=1.4\hsize\raggedright\arraybackslash}X}
  \begin{tabularx}{\textwidth}{@{}l Y Z@{}}
    \toprule
    \textbf{Layer} & \textbf{Key components}
                   & \textbf{Architectural responsibility} \\
    \midrule
    \textbf{Data}
    & Data lakehouse; AI-ready store; results database;
      model artifact catalog
    & Owns the single source of truth: guarantees that raw
      imagery, the analysis-ready representation, extracted
      traits, and model versions are persistent, consistently
      indexed, and reproducibly addressable across queries. \\
    \addlinespace
    \textbf{Compute}
    & AI-readiness pipeline; ViT trait-extraction service on
      Frontier; Parsl orchestration
    & Owns scale: converts heterogeneous sensor data into
      AI-ready form and turns trait requests into parallel
      inference on Frontier, decoupling preparation cost from
      analysis cost. \\
    \addlinespace
    \textbf{Agent}
    & Co-Scientist Agent; Compute Agent; S3M secure channel
    & Owns autonomy and federation: translates intent into
      executable plans, decides what must be (re)computed, and
      coordinates across two security domains without
      persistent coupling. \\
    \addlinespace
    \textbf{UI}
    & Chainlit web chat; report and visualization generator
    & Owns the human interface: captures scientific intent in
      natural language and returns interpretable visual and
      textual results that close the loop with the scientist. \\
    \bottomrule
  \end{tabularx}
\end{table*}

\begin{compactenum}
  \item \textbf{Context specification.} The scientist specifies the analysis context in natural language: the date range of interest, the imaging modality or camera, the trait(s) to examine, and the criterion by   which results should be ranked (e.g., which poplar genotypes showed the greatest height gain over the last two weeks?).
  
  \item \textbf{Availability check and on-demand inference.} The system resolves the request against the results database and checks whether the requested traits are already available for the specified context. If so, they are served directly; if they are missing or stale, the agentic layer triggers new trait-extraction jobs on Frontier and waits for them to complete, so that the scientist never has to know whether a result was precomputed or generated on-demand.
  
  \item \textbf{Aggregation, analysis, and result generation.} Once the required traits are available, the agents aggregate them across plants and time, apply the requested ranking or comparison, and generate both visual artifacts (segmentation overlays and trait time-series plots) and a textual interpretation that summarizes the biological finding.
  
  \item \textbf{Follow-up with context reuse.} The scientist asks follow-up questions within the same session. Because the established context and any  computed traits are retained, refinements such as narrowing the date range, changing the ranking criterion, or drilling into a single genotype are answered quickly, without recomputing work that has already been done.
\end{compactenum}

This loop is what turns a high-throughput data factory into an interactive discovery platform: the first question in a line of inquiry may incur inference costs, but subsequent questions are typically answered in seconds.

While the loop above describes the pipeline behaviorally, as the sequence of actions a query triggers, Table~\ref{tab:layers} describes it structurally, decomposing the same system into four layers and the distinct responsibilities each one holds.

\section{From Multimodal Images to Time-resolved Plant Traits}
\label{sec:images_to_traits}

\begin{figure}[!ht]
    \centering
    \includegraphics[width=\linewidth]{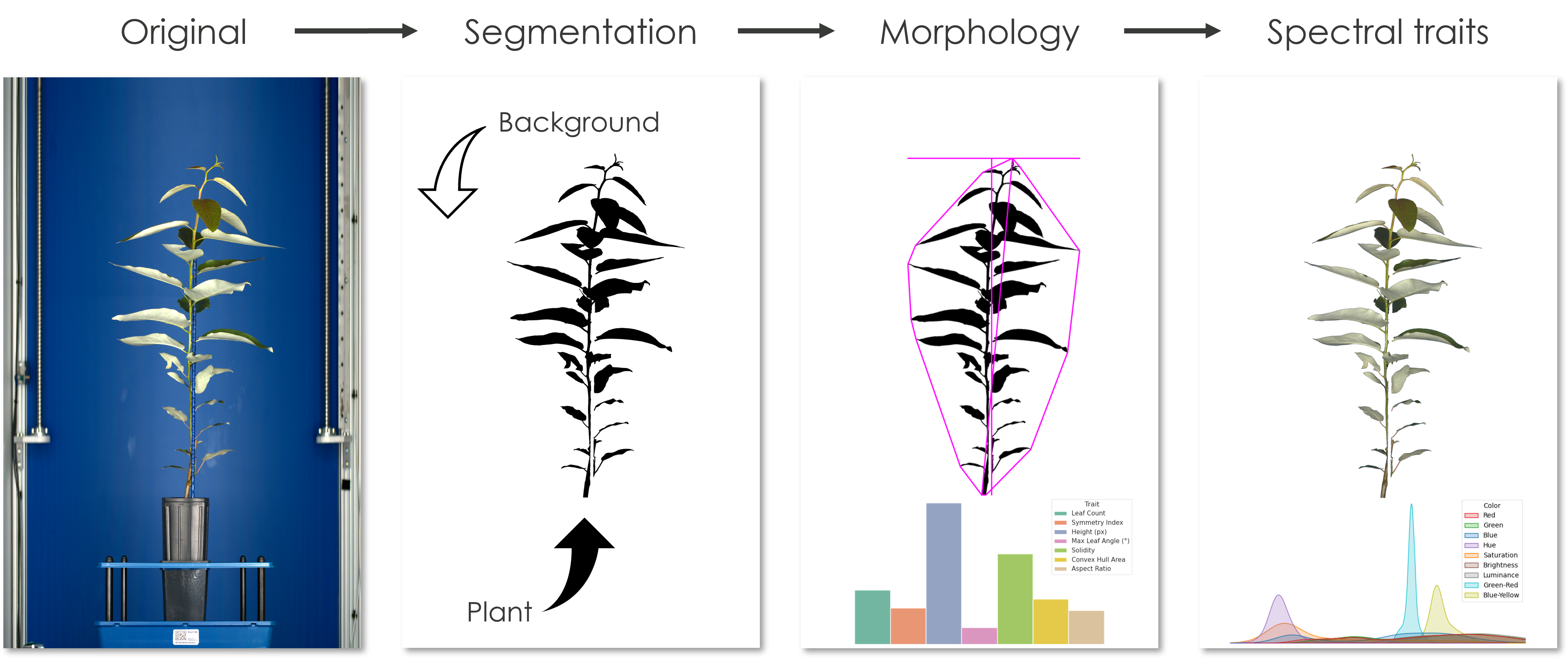}
    \caption{From a raw image to traits. A representative RGB plant image is segmented into a plant/background mask, from which morphological traits (for example, height, projected leaf area, and shape indices) are computed directly and, together with the spectral and thermal modalities, physiological proxies (for example, $F_v/F_m$ and leaf temperature) are derived.}
    \label{fig:images_to_traits}
\end{figure}

Turning APPL's complementary raw imaging modalities into time-resolved, comparable plant traits requires three stages, shown in Fig.~\ref{fig:images_to_traits}: making heterogeneous sensor data AI-ready, segmenting the plant from the background at full resolution, and reducing each image/mask pair into quantitative traits. The first stage (Section~\ref{sec:airready}) standardizes the modality-specific formats produced by the conveyor's cameras into a common, metadata-linked representation. The second stage (Section~\ref{sec:ViT}) applies a ViT-based segmentation model, deployed on Frontier, to isolate plant pixels under wide natural variations in size, shape, and developmental stage. The third stage (Section~\ref{sec:traitextraction}) aggregates pixel-level measurements over the segmented region into one scalar per trait per plant per imaging round, yielding the tidy, analysis-ready dataset that the agentic framework reasons over.

\subsection{Making Imaging Data AI-Ready}
\label{sec:airready}

Each of the imaging modalities captured by the APPL phenotyping platform requires dedicated preprocessing before any computational analysis can proceed. Raw images are loaded via modality-specific readers and converted to standardized NumPy arrays using Python (3.12), decoupling sensor-level specifics from downstream processing.

Two RGB cameras yield side-view (RGB1) and top-down (RGB2) imagery stored as three-channel arrays. A chlorophyll fluorescence imager (FC1) produces binary files encoding four channels per pixel (i.e., minimal fluorescence $F_o$, maximal fluorescence $F_m$, variable fluorescence $F_v$, and the photochemical efficiency ratio $F_v/F_m$). An infrared camera (IR1) delivers raw pixel intensities mapped through a hardware lookup table to surface temperatures (Kelvin, converted to Celsius), yielding a single-channel image. Finally, two push-broom hyperspectral cameras record reflectance across the visible and near-infrared spectrum: VNIR (480 bands, 380--900 nm) and SWIR (636 bands, 900--1700 nm), both stored in Band Interleaved by Line (BIL) format and normalized against per-session white and dark references.

All modalities share a common metadata schema, linking each image to an experiment identifier (one of 70+ tracked experiments spanning poplar, switchgrass, pennycress, and \textit{Arabidopsis thaliana} accessions), a plant identifier (encoding genotype and tray position), a round identifier (the imaging session, enabling longitudinal tracking), and a modality tag. NumPy file names follow the convention \verb+{experiment_id}__{plant_id}__{round_id}+ to ensure that metadata and image arrays remain unambiguously associated throughout the pipeline without a separate database lookup. In images with multiple plants per tray, the first plant ID is used as a reference for all other plants in the tray to avoid duplication. Per-plant analysis is performed by cropping to each plant's associated region of interest (ROI).

\subsection{Vision Transformer-base Image Segmentation}
\label{sec:ViT}

Quantitative trait extraction requires isolating plant pixels from background (i.e., pots, soil, support structures, and imaging artifacts). This is a non-trivial task at high resolution and under natural variation in plant size, shape, and developmental stage. Classical thresholding and hand-crafted feature detectors fail to generalize across growth stages or imaging conditions, motivating the use of a learned segmentation model.

We adopt a Vision Transformer (ViT)~\cite{ViT} as the encoder backbone, paired with a lightweight convolutional decoder. Compared to convolutional networks, ViTs capture long-range spatial dependencies through self-attention, which is particularly beneficial for segmenting complex, non-convex plant shapes that span a large portion of the image. Recent high-throughput phenotyping pipelines have explored alternative approaches, including YOLO-family detectors and promptable models such as the Segment Anything Model (SAM)~\cite{wang2025sensors}; these offer attractive zero- or low-shot capabilities but tend to degrade on small, occluded, or fine-edge structures and either depend on accurate prompts or struggle to generalize across imaging modalities and developmental stages. A fine-tuned ViT trained directly on APPL imagery sidesteps both limitations and, critically, exposes intermediate features that can later be repurposed as a perceptual backbone for OPAL's biology foundation-model effort. The encoder is instantiated as a \verb+vit_base+ model with a patch size of 8 and a window size of 448, pretrained on ImageNet, and fine-tuned on annotated APPL imagery. The decoder applies successive upsampling blocks with residual connections to recover full-resolution binary segmentation maps (plant vs.~background).

\begin{figure}[!h]
    \centering
    \includegraphics[width=\linewidth]{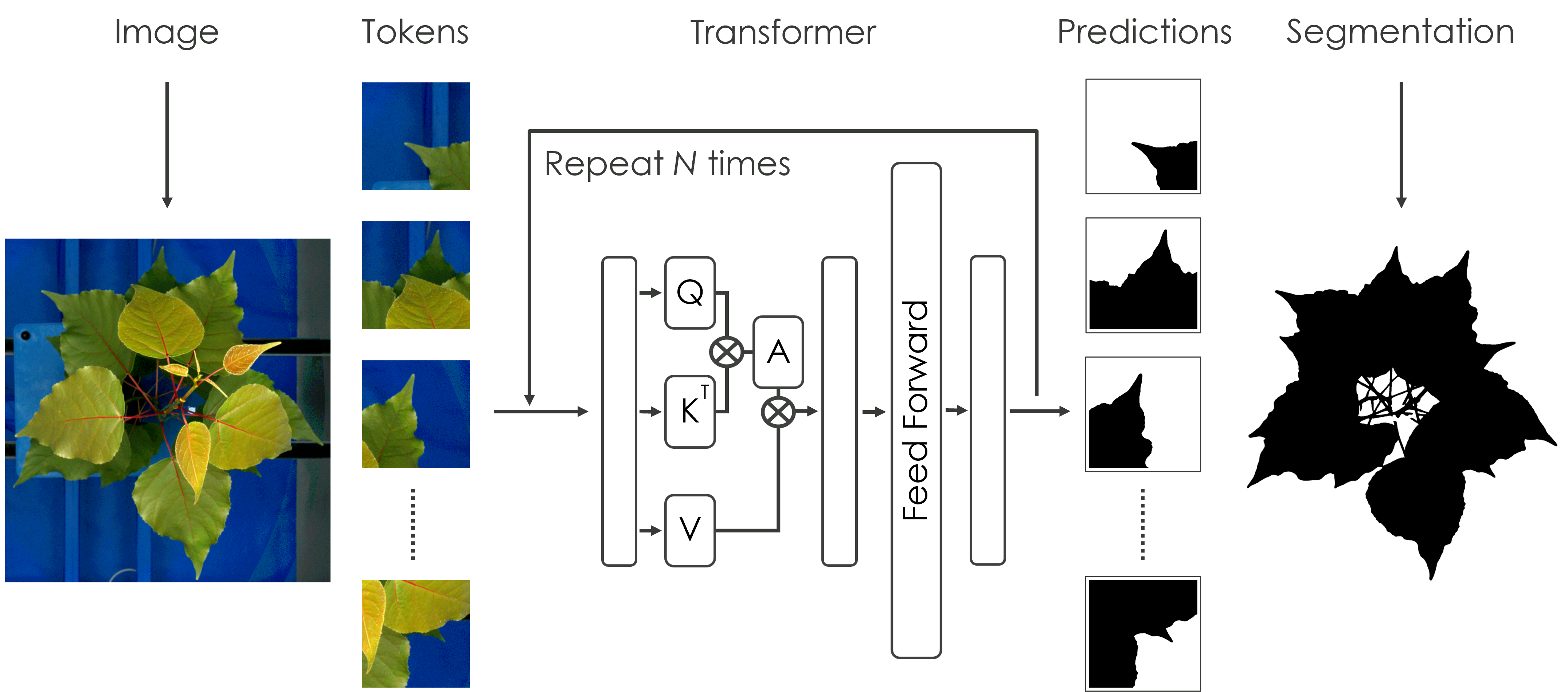}
    \caption{ViT-based segmentation: a high-resolution input is split into overlapping tiles, each encoded by a Vision Transformer and decoded into a per-tile mask, which are blended into a full-image plant/background segmentation.}
    \label{fig:ViT-segmentation}
\end{figure}

Because the raw images are significantly larger than the ViT's input window (448 pixels), inference proceeds via a sliding-window strategy. Overlapping tiles are extracted with a 50\% stride and normalized to ImageNet statistics. Each tile is passed through the model in batches, producing per-class probability maps that are blended back into a full-image canvas using Hann-window weighting to suppress boundary artifacts at tile edges. The final binary mask is obtained by taking the $\arg\max$ over the accumulated probability map. To eliminate spurious isolated predictions unconnected to the main plant body, a graph-based pruning step identifies connected components, clusters them by axis-aligned bounding-box overlap, and retains only the largest spatially coherent cluster. Fig.~\ref{fig:ViT-segmentation} illustrates the full process, from tiled high-resolution input through per-tile ViT encoding and decoding to the final blended segmentation mask.

\begin{figure*}[!ht]
    \centering
    \includegraphics[width=.94\textwidth]{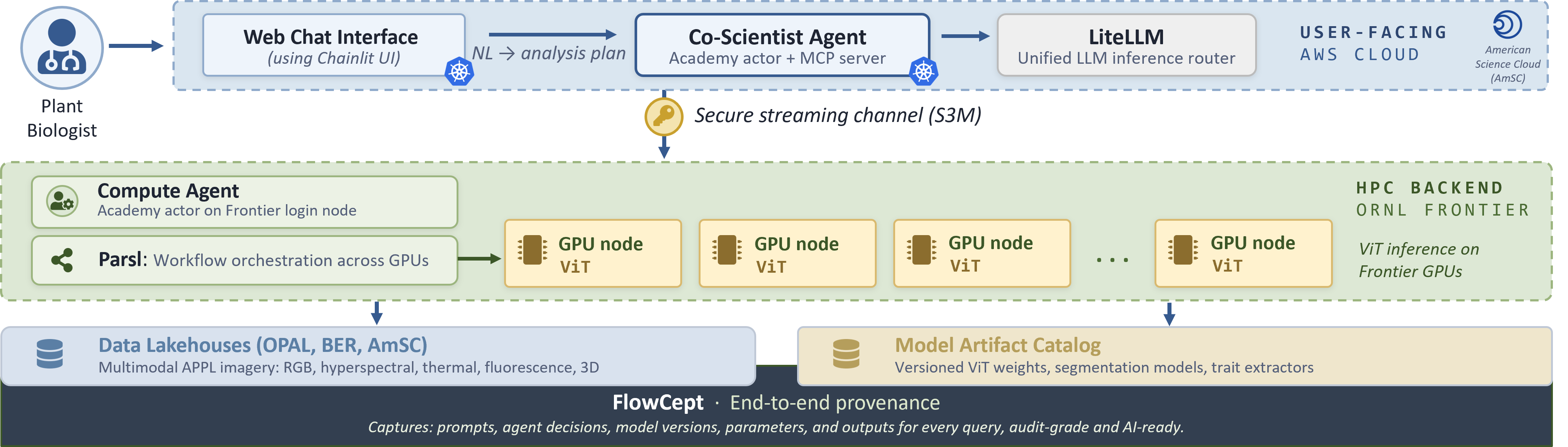}
    \caption{End-to-end architecture of the APPL agentic framework. Two execution domains are bridged by S3M, a secure, token-authenticated streaming channel. The user-facing zone exposes a Chainlit web chat interface backed by a Co-Scientist Agent and a LiteLLM inference router. The HPC backend zone, hosted on Frontier, runs a Compute Agent that dispatches ViT segmentation and trait extraction across GPU nodes via Parsl. Imagery is read from the OPAL data lakehouse and model weights from a versioned artifact catalog. FlowCept captures every interaction from natural-language prompt to returned result for auditability and reproducibility.}
    \label{fig:deployment}
\end{figure*}

\subsection{Time-Resolved Plant Traits Extraction}
\label{sec:traitextraction}

Once per-image segmentation masks are available, quantitative traits are extracted for each plant at each imaging round, enabling longitudinal phenotyping across the full experiment. 

Morphological traits are computed directly from the binary mask on RGB images. Plant height is measured from the bottom of the image to the topmost plant pixel in the RGB1 side view and is converted to millimeters using the calibrated pixel scale. Projected leaf area (in mm$^2$) is computed from the RGB2 top-down mask area. Plant width is estimated from the bounding-box width of the pruned mask in the side view. Shape indices (i.e., compactness and roundness) characterize the space-filling geometry of the canopy. From the FC1 fluorescence modality, the ratio $F_v/F_m$ provides a per-pixel estimate of the maximum quantum yield of photosystem II, serving as a proxy for photosynthetic efficiency and early stress detection. Leaf temperature derived from IR1 thermal imagery is a proxy for stomatal conductance, as transpiration cools the leaf surface. Chlorophyll content is approximated from RGB color features (green-to-red ratio in CIE Lab space) and from the $F_v/F_m$ signal.

All pixel-level measurements are aggregated over the segmented plant region using the median (or mean), yielding a single scalar per trait per plant per round. Results are stored as per-sample CSV files keyed on (experiment\_id, plant\_id, round\_id) and subsequently merged into a consolidated modality-level table (e.g., rgb1\_features.csv) enriched with the full APPL metadata (genotype, treatment, replicate, date, time, etc.), producing a tidy, analysis-ready dataset for downstream statistical modeling and agentic reasoning.

Several ecophysiologically important traits are not yet derived from imaging alone: photosynthetic rate and stomatal conductance (requiring gas-exchange measurements or calibrated spectral models), aboveground biomass and leaf mass per area (currently requiring destructive sampling, though allometric models from 3D reconstructions are planned), total protein content, and volatile emissions (isoprene, methanol) measured by mass spectrometry. Integration of these traits into the agentic framework will rely on fine-tuning multimodal foundation models for APPL, currently under active development within the OPAL project.

\section{The Agentic Framework: Architecture, Implementation, and Deployment}
\label{sec:implem_deploy}

Fig.~\ref{fig:deployment} details how the logical pipeline of Section~\ref{sec:system}, and specifically the Agent layer of Table~\ref{tab:layers}, is realized as two execution domains that mirror the operational reality of running an agentic system inside a DOE national laboratory. A user-facing zone, hosted as Docker containers on Amazon Elastic Kubernetes Service, is where a plant biologist poses natural-language questions through a Chainlit web chat interface, interpreted by a Co-Scientist Agent backed by a LiteLLM routing service. An HPC backend zone, hosted on the Frontier exascale supercomputer inside Oak Ridge National Laboratory's secure perimeter, is where a Compute Agent orchestrates ViT inference across GPU nodes via Parsl~\cite{parsl}. The two zones are bridged by S3M~\cite{s3m}, a secure, token-authenticated streaming channel, with FlowCept~\cite{flowcept} capturing end-to-end provenance across both. The remainder of this section describes each element in turn: Section~\ref{sec:agentic-framework} motivates the two-zone design and our choice of agentic framework; Section~\ref{sec:copilot} details the Co-Scientist Agent; Section~\ref{sec:compute-agent} describes the Compute Agent and its Parsl-based orchestration; and Section~\ref{sec:a2a} presents the S3M communication channel.

\subsection{Agentic Framework}
\label{sec:agentic-framework}
Scientific workflows naturally decompose into interacting stages, such as simulation, analysis, steering, and experimental control, each with distinct computational demands and resource requirements. Intelligent agents are well suited to encapsulating these stages: each maintains its own persistent state, exposes well-defined behaviors, and can delegate to or synthesize the outputs of peer agents. When composed into multi-agent systems, this modularity enables specialized agents to coordinate dynamically across complex, multi-step campaigns that no single agent could manage alone, while remaining individually replaceable and extensible.

Despite this promise, existing agentic frameworks fall short of what scientific applications require. Popular frameworks such as LangGraph~\cite{langgraph}, AutoGen~\cite{autogen}, and CrewAI~\cite{crewai2026} are built primarily for conversational, cloud-native use cases, adopting centralized architectures with a single orchestrator, cloud-bound message buses, and tight coupling between control and data planes. They are poorly matched to the federated, heterogeneous, and often intermittently connected infrastructure, the multi-gigabyte data movements between instruments and HPC nodes, and the long-running stateful agents that characterize real research environments. Using an off-the-shelf agentic framework on a leadership-class supercomputer is, in practice, neither straightforward nor wise: scientific workflows demand asynchronous execution against dynamic resource availability, separation of control and data planes, and provenance-aware coordination that conversational frameworks were not designed to provide.

Academy~\cite{academy} addresses these challenges through a modular and extensible middleware designed specifically for deploying autonomous agents across federated research infrastructures, including HPC systems, experimental facilities, and data repositories. Rather than adapting a general-purpose conversational framework to the demands of science, Academy was designed from the ground up with the realities of research computing in mind. Its architecture is built around the actor model, a concurrent computing paradigm in which agents encapsulate local state and communicate through message passing, extended with the ability of each agent to engage independently with its environment and with other agents.

At a high level, Academy organizes agents into a federated ecosystem where each agent manages its own state and exposes well-defined behaviors to peer agents or human users. The framework provides flexible communication abstractions that accommodate both low-latency local interactions and high-throughput data transfers across facilities, and it supports asynchronous execution to ensure that agents remain productive even when remote resources are temporarily unavailable. Agents can be deployed on HPC clusters, experimental endpoints, or cloud nodes, and the framework handles the underlying coordination transparently.

\begin{figure}[!ht]
    \centering
    \includegraphics[width=\linewidth]{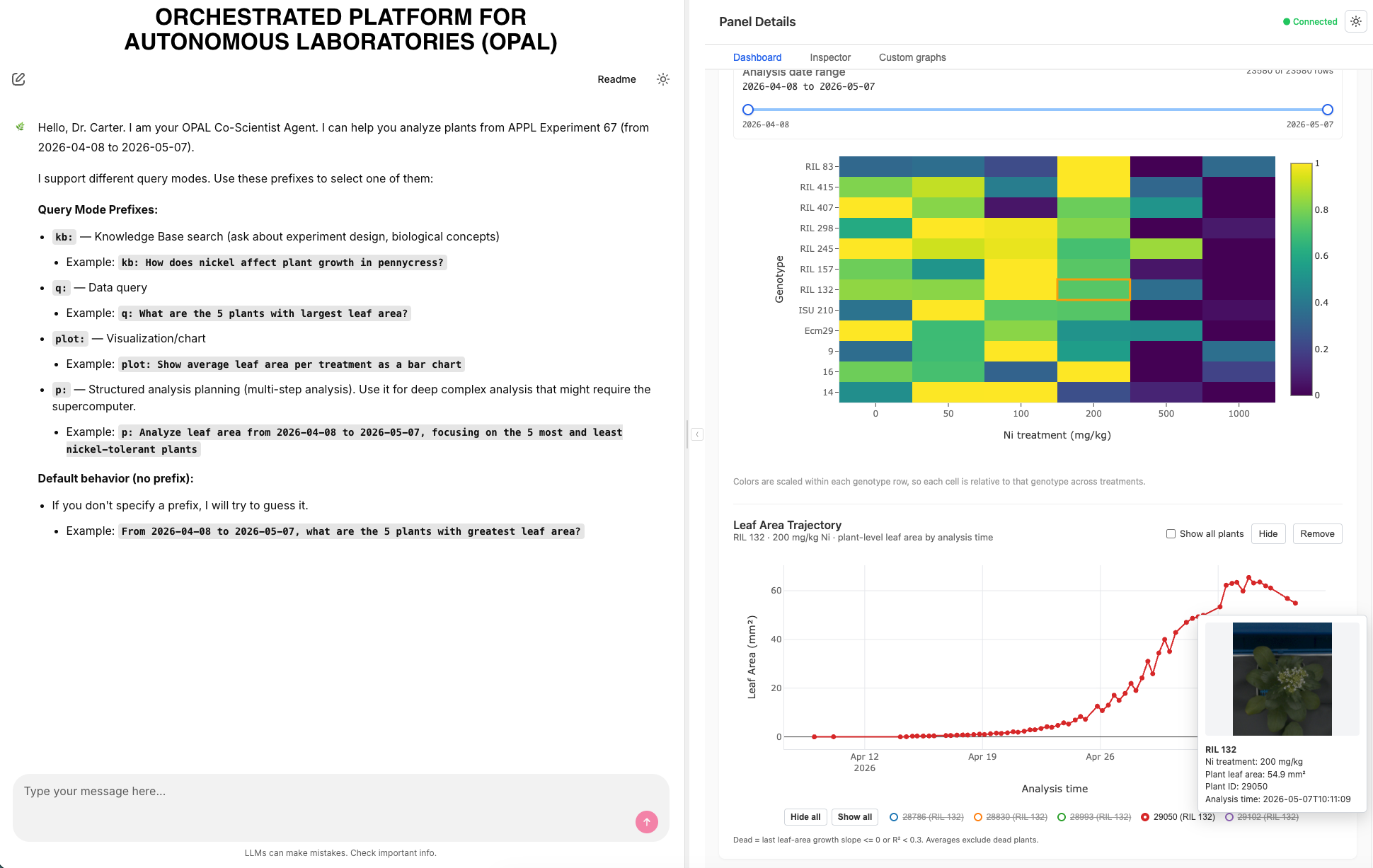}
    \caption{Co-Scientist agent Chainlit-based chat\,+\,dashboard user interface.}
    \label{fig:chainlit-ui}
\end{figure}

\subsection{Co-Scientist Agent}
\label{sec:copilot}

The APPL Co-Scientist Agent is the conversational front end of the agentic system, designed to guide plant biologists through the specification and execution of image analysis workflows via natural language. It is implemented as an MCP server backed by an Academy agent and exposed to users through a Chainlit web interface illustrated in Fig.~\ref{fig:chainlit-ui}.

On startup, the agent greets the user with a personalized message summarizing the active experiment and its temporal coverage, then enters a structured requirements-gathering dialog. Driven by a detailed system prompt, the underlying LLM steers the conversation through four mandatory plan components: date range, imaging modality, plant traits of interest, and a plant selection strategy.

The selection strategy supports uniform sampling, explicit plant enumeration, and score-ranked modes (top, worst, or top-and-worst performers) based on user-defined trait formulae. Throughout the dialog, the agent resolves natural-language expressions (e.g., "last two weeks", "tallest plants") into structured values, maps common trait names to their internal column identifiers, and answers incidental biology questions without losing conversational context. 

Users may also issue ad hoc analytical queries against experiment metadata at any time, such as execution times, CPU utilization, or metadata attributes. Once the user signals completion, the Co-Scientist serializes the collected parameters into a structured JSON plan and delegates execution to the APPL Compute Agent via an Academy remote action call. Upon receiving results, it extracts trait time-series data from the returned figures and invokes the LLM once more to produce a structured Markdown biological report entitled ``APPL Analysis Report'', comparing winner and loser plants, interpreting growth dynamics, and recommending follow-up analyses. Upon a validation request, the agent can additionally generate a ``Provenance-based Report Verification''. Results are rendered in the Chainlit interface alongside segmentation visualizations and trait plots.

\subsection{Compute Agent}
\label{sec:compute-agent}

The APPL Compute Agent is a headless, GPU-oriented backend that operates as an Academy agent, accepting structured analysis plans from the Co-Scientist and relieving the Co-Scientist of heavy inference workloads.

Upon receiving a plan, the compute agent selects plants from experiment metadata according to the user-specified criteria (random sampling, explicit enumeration, or performance-ranked selection) and retrieves the corresponding multispectral image files (RGB, infrared, or fluorescence modalities). Segmentation inference is then performed using the pre-trained ViT model introduced in Section~\ref{sec:ViT}, with Parsl~\cite{parsl} managing parallel task dispatch across GPU nodes on the Frontier leadership class supercomputer. Cached segmentation masks are reused to avoid redundant computation.

Once segmentation is complete, the agent extracts plant traits, ranks plants as winners or losers relative to the scoring formula, and produces diagnostic figures showing raw images, segmentation masks, and trait time series. Workflow provenance (campaign identifiers, input parameters, and output summaries) is captured via FlowCept~\cite{flowcept} to support reproducibility and experimental traceability.

\subsection{Agent-to-Agent Communication}
\label{sec:a2a}


The Co-Scientist and Compute agents run in different security and resource domains: the Co-Scientist is exposed to end users via a Chainlit UI in an EKS pod on AWS, while the Compute Agent operates on Frontier login and compute nodes inside ORNL's secure perimeter. Direct, persistent network coupling between these two environments would be both impractical and undesirable. We therefore mediate all agent-to-agent communication through S3M~\cite{s3m}, a secure, API-driven, token-authenticated streaming service developed for cross-facility research automation. S3M provides authenticated, audit-logged channels for delivering structured JSON analysis plans, streaming intermediate progress, and returning result artifacts, allowing the Co-Scientist Agent to invoke remote Academy actions on the Compute Agent as if they were local, while preserving the institutional security boundaries that real DOE deployments require. Every S3M-mediated exchange is captured by FlowCept, producing a complete, auditable trail from natural-language query to delivered scientific insight.

\subsection{Deployment Experience}
\label{sec:experience}

The framework has been in use against APPL's live data lakehouse since late 2025, serving biologists across experiments spanning poplar, switchgrass, pennycress, and Arabidopsis. For a representative query (top- and worst-performing genotypes by height gain over a two-week window, RGB1 modality, 498 plants), a cold request that triggers segmentation on Frontier completes in about 5 minutes for eight plants; once traits are cached, semantically related follow-up queries return in less than a minute, and end-to-end provenance captured by FlowCept adds less than 1\% execution time overhead. These figures are illustrative of the interactive regime the framework targets rather than a controlled benchmark. More concretely, the agentic workflow has transformed what was previously a manual process requiring approximately two hours per day over six days of a forty-day experiment into a fully automated pipeline that completes within minutes of new images being acquired. Extracted traits are immediately available for reasoning, reducing analysis latency from a multi-day manual workflow to an automated, query-driven process. 

\section{Conclusion and Future Work}
\label{sec:ccl}

We presented an agentic AI framework that turns a high-throughput plant phenotyping facility from a data factory into an interactive discovery platform. The contribution is not a single model but an integration: HPC-scale Vision-Transformer perception, a federated multi-agent architecture that separates a cloud-hosted conversational plane from an HPC execution plane, a secure cross-domain channel between them, and end-to-end provenance over every interaction. Deployed in production against APPL's live data lakehouse, the framework collapses a days-to-weeks analysis cycle into a loop in which follow-up questions are answered in seconds.

For the eScience community, the broader lesson is that bringing agentic AI to leadership-class science is less a modeling problem than an infrastructure one: the hard parts are federation across security domains, separation of control and data planes, and provenance rigorous enough to be reused as training data. Several directions follow. We will extend trait extraction to the hyperspectral and 3D modalities and fuse imaging-derived proxies with co-located gas-exchange and mass-spectrometry streams; we will harden the provenance corpus into curated training data for biology foundation models; and we will extend the Academy deployment beyond ORNL to coordinate experiments across the OPAL national laboratories, a step toward the federated autonomous-science network the DOE Genesis Mission envisions.
\section*{Acknowledgments}

This material by the Orchestrated Platform for Autonomous Laboratories (OPAL) is based upon work supported by the U.S. Department of Energy, Office of Science, through the Office of Biological and Environmental Research Program, under contracts DE-AC02-06CH11357 (ANL); DE-AC02-05CH11231 (LBNL); DE-AC05-00OR22725 (ORNL); and DE-AC05-76RL01830 (PNNL). This material is also based upon work at the Center for Bioenergy Innovation supported by the U.S. Department of Energy, Office of Science, Biological and Environmental Research under Contract number ERKP886. This research used resources of the Oak Ridge Leadership Computing Facility and the Advanced Plant Phenotyping Laboratory at the Oak Ridge National Laboratory, supported by the Office of Science of the U.S. Department of Energy under Contract No. DE-AC05-00OR22725.

\bibliographystyle{IEEEtran}
\bibliography{references}

\end{document}